\documentclass[letterpaper, 10 pt, conference]{ieeeconf}  

\IEEEoverridecommandlockouts                              

\usepackage{comment}
\usepackage{cite}
\usepackage{amsmath,amssymb,amsfonts}
\usepackage{algorithmic}
\usepackage{graphicx}
\usepackage{subcaption}
\usepackage{tabularray}
\usepackage{caption}

\usepackage[dvipsnames,table]{xcolor}
\usepackage{tikz}
\usepackage{academicons}
\definecolor{orcidlogocol}{HTML}{A6CE39}

\usepackage{tabularx}
\usepackage{nicematrix}

\newcolumntype{x}[1]{!{\centering\arraybackslash\vrule width #1}}
\newcolumntype{L}[1]{>{\raggedright\arraybackslash}p{#1}} 

\usepackage{array}
\usepackage{multirow}

\newcolumntype{R}[2]{%
    >{\adjustbox{angle=#1,lap=\width-(#2)}\bgroup}%
    l%
    <{\egroup}%
}

\usepackage{arydshln}

\usepackage{booktabs}
\usepackage{chngcntr} 
\counterwithout{figure}{section} 
\usepackage[export]{adjustbox}

\usepackage[inkscapelatex=false]{svg}

\definecolor{lime}{HTML}{A6CE39}

\DeclareRobustCommand{\orcidicon}{%
    \begin{tikzpicture}
    \draw[lime, fill=lime] (0,0) 
    circle [radius=0.16] 
    node[white] {{\fontfamily{qag}\selectfont \tiny ID}};
    \draw[white, fill=white] (-0.0625,0.095) 
    circle [radius=0.007];
    \end{tikzpicture}
    \hspace{-2mm}
}

\newcommand{\orcidWalter}{\href{https://orcid.org/0000-0003-4565-1272}{\orcidicon}}
\newcommand{\orcidChristian}{\href{https://orcid.org/0000-0003-4822-2844}{\orcidicon}}
\newcommand{\orcidRoss}{\href{https://orcid.org/0000-0001-8595-0379}{\orcidicon}}
\newcommand{\orcidAhmed}{\href{https://orcid.org/0000-0003-3702-8042}{\orcidicon}}
\newcommand{\orcidBjoerk}{\href{https://orcid.org/0009-0000-6686-0394}{\orcidicon}}
\newcommand{\orcidAndreas}{\href{https://orcid.org/0000-0003-0328-382X}{\orcidicon}}
\newcommand{\orcidTrivedi}{\href{https://orcid.org/0000-0002-0937-6771}{\orcidicon}}
\newcommand{\orcidKnoll
}{\href{https://orcid.org/0000-0003-4840-076X}{\orcidicon}}
\usepackage{textcomp}

\usepackage{siunitx}
\def\BibTeX{{\rm B\kern-.05em{\sc i\kern-.025em b}\kern-.08em
    T\kern-.1667em\lower.7ex\hbox{E}\kern-.125emX}}

\makeatletter 
\newcommand{\linebreakand}{%
  \end{@IEEEauthorhalign}
  \hfill\mbox{}\par
  \mbox{}\hfill\begin{@IEEEauthorhalign}
}    
    
\makeatletter
\newcommand*{\emails}[2][@tum.de]{%
    \def\@tempa{\@gobble}%
    \@for\qrr@email:=#2\do{%
        \edef\@tempb{\noexpand\href{mailto:\qrr@email #1}{\qrr@email}}%
        \edef\@tempa{\unexpanded\expandafter{\@tempa}{, }\unexpanded\expandafter{\@tempb}}}%
    \{\@tempa\}#1%
}    

\usepackage{url}
\usepackage{capt-of}
\usepackage{etoolbox}
\usepackage{wasysym}

\makeatletter
\let\NAT@parse\undefined
\makeatother
\usepackage{hyperref} 
\hypersetup{
    bookmarks=false,
    colorlinks=true,
    breaklinks=true,
    linkcolor=blue,
    filecolor=magenta,      
    hypertexnames=true,
    urlcolor=cyan,
    pagebackref=true,
    pdftitle={ITSC2023},
    pdfpagemode=FullScreen,
    citecolor=green
    }
\usepackage[capitalize]{cleveref}

\crefname{figure}{Fig.}{Figs.}
\Crefname{figure}{Figure}{Figures}
\crefname{section}{Sec.}{Secs.}
\Crefname{section}{Section}{Sections}
\crefname{table}{Tab.}{Tabs.}
\Crefname{table}{Table}{Tables}
\crefname{equation}{Eq.}{Eqs.}
\Crefname{equation}{Equation}{Equations}

\title{\LARGE \bf
ActiveAnno3D - An Active Learning Framework for\\ Multi-Modal 3D Object Detection
}

\author{Ahmed Ghita$^{1*}$\orcidAhmed, Bj\o rk Antoniussen$^{2,3*}$\orcidBjoerk, Walter Zimmer$^{1*}$\orcidWalter, Ross Greer$^{2*}$\orcidRoss, Christian Creß$^{1*}$\orcidChristian, \\Andreas M\o gelmose$^{3}$\orcidAndreas, Mohan M. Trivedi$^{2}$\orcidTrivedi, and Alois C. Knoll$^{1}$\orcidKnoll
\thanks{$^{1}$A. Ghita, W. Zimmer, C. Creß, and A. Knoll are with the Chair of Robotics and Artificial Intelligence at the Technical University of Munich. Corresponding author: W. Zimmer
        {\tt\small walter.zimmer@tum.de}}%
\thanks{$^{2}$B. Antoniussen, R. Greer, and M. Trivedi are with the Laboratory for Intelligent and Safe Automobiles at the University of California San Diego.}
\thanks{$^{3}$B. Antoniussen and A. M\o gelmose are with the Visual Analysis and Perception Lab at Aalborg Universitet.}
\thanks{$^{*}$Authors contributed equally.}
}

\begin{document}

\makeatletter
\let\@oldmaketitle\@maketitle
\renewcommand{\@maketitle}{\@oldmaketitle
    \centering

  \includegraphics[width=\linewidth]{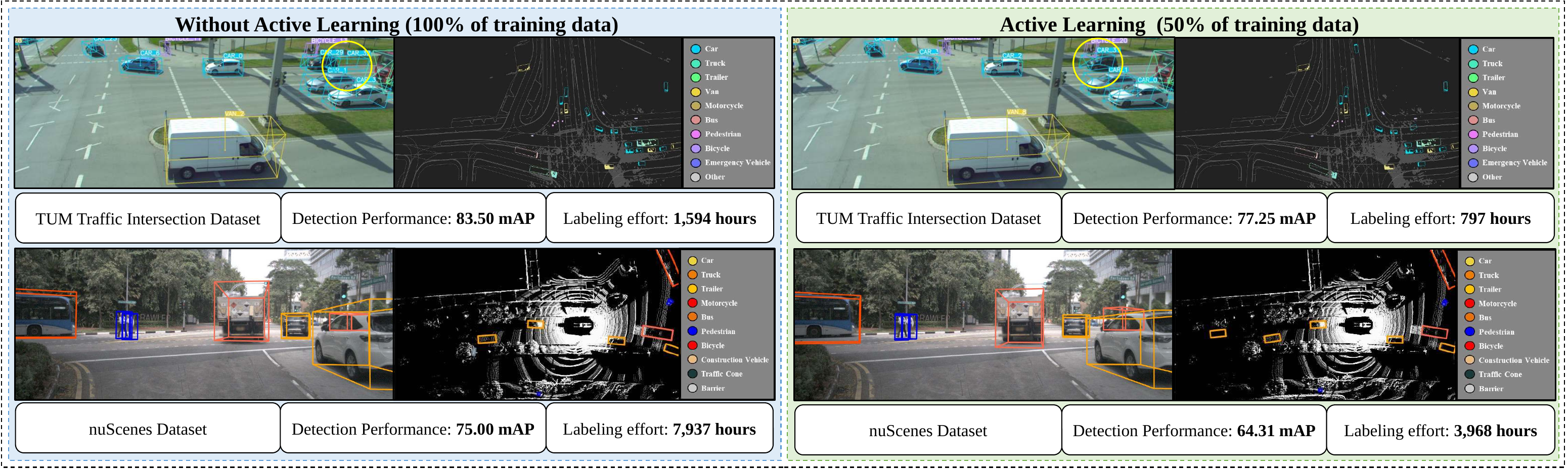}\bigskip
  \captionof{figure}{We propose a framework for efficient active learning within various 3D object detection techniques and modalities, demonstrating the effectiveness of active learning at reaching comparable detection performance on benchmark datasets at a fraction of the annotation cost. Datasets include roadside infrastructure sensors (top row) and onboard vehicle sensors (bottom row), with LiDAR-only and LiDAR+camera fusion methods, the two dominant strategies in state-of-the-art performance at the safety-critical detection task.}
  \label{fig:overview_figure}
  }
  
\makeatother

\maketitle
\thispagestyle{empty}
\pagestyle{empty}

\begin{abstract}

The curation of large-scale datasets is still costly and requires much time and resources. Data is often manually labeled, and the challenge of creating high-quality datasets remains. In this work, we fill the research gap using active learning for multi-modal 3D object detection. We propose \textit{ActiveAnno3D}, an active learning framework to select data samples for labeling that are of maximum informativeness for training. We explore various continuous training methods and integrate the most efficient method regarding computational demand and detection performance. Furthermore, we perform extensive experiments and ablation studies with BEVFusion and PV-RCNN on the nuScenes and TUM Traffic Intersection dataset. We show that we can achieve almost the same performance with PV-RCNN and the entropy-based query strategy when using only half of the training data (77.25 mAP compared to 83.50 mAP) of the TUM Traffic Intersection dataset. BEVFusion achieved an mAP of 64.31 when using half of the training data and 75.0 mAP when using the complete nuScenes dataset. We integrate our active learning framework into the proAnno labeling tool to enable AI-assisted data selection and labeling and minimize the labeling costs. Finally, we provide code, weights, and visualization results on our website: \url{https://active3d-framework.github.io/active3d-framework}.

\end{abstract}


\newcolumntype{N}{@{}m{0pt}@{}}
\begin{table*}[t]

    \caption{Comparison of previous research in Active Learning for 3D Object Detection in Autonomous Driving Datasets.}
    \centering
    \begin{tabular*}{\textwidth}{p{6.5cm}p{.7cm}p{1.6cm}p{2.1cm}p{4.8cm}}
        \hline
        \textbf{Active Learning Methods} & \textbf{Year} & \textbf{Datasets} & \textbf{Modalities} & \textbf{Limitations} \\
        \hline \hline 
        Entropy, Monte Carlo dropout, ensemble learning \cite{feng2019deep} & 2019 & KITTI & LiDAR, Camera$^\dag$ & Image detector excluded from active learning loop \\ \hline 
        Uncertainty sampling \cite{meyer2019automotive} & 2019 & Astyx \cite{meyer2019automotive} & Radar, Camera, LiDAR & Small dataset, little environmental variance \\ \hline 
        Semi-supervised co-training on prediction disagreement \cite{villalonga2020co} & 2020 & KITTI, Waymo & Camera & Single-mode approaches do not provide necessary  accuracy for domain \\ \hline
        Consensus score variation ratio, sequential region-of-interest matching \cite{schmidt2020advanced} & 2020 & KITTI & LiDAR+Camera & Hardware limitations prevent training backbone to convergence \\ \hline 
        Bayesian surprise (KL divergence) \cite{ccatal2020anomaly}  & 2020 & AGV Anomaly Dataset & Camera & Single-mode approaches do not provide necessary  accuracy for domain \\ \hline 
        Augmentation, dropout, insertion, deletion \cite{meng2021towards} &2021 & KITTI & LiDAR & Sparse LiDAR points necessitate camera; pedestrians are challenging \\ \hline  
        Class Entropy and Spatial Uncertainty \cite{moses2022localization} & 2022 & Private & LiDAR & Single-mode approaches do not provide necessary  accuracy for domain \\ \hline 
        Uncertainty sampling \cite{chen2022gocomfort} & 2022 & Private & LiDAR+Camera & Detection of road damage; does not address object collision \\ \hline 
        
        Spatial and temporal diversity-based sampling \cite{liang2022exploring} & 2022 & nuScenes & LiDAR & Model uncertainty may improve AL for task performance; single sensing modality \\ \hline
        Kernel coding rate \cite{luo2023kecor} & 2023 & KITTI, Waymo & LiDAR & Single-mode approaches do not provide necessary  accuracy for domain \\ \hline 
        3D consistency of bounding box predictions in both semi-supervised and active learning \cite{hwang2023joint} & 2023 & KITTI & LiDAR & Plans to evaluate on nuScenes for robustness \\ \hline 
        Bi-domain active learning, diversity-based sampling  \cite{yuan2023bi3d} & 2023 & KITTI, nuScenes, Waymo, Lyft & LiDAR & Single-mode approaches do not provide necessary  accuracy for domain \\ \hline  
        Ego-pose distance-based sampling \cite{almin2023navya3dseg} & 2023 & Navya3DSeg & LiDAR & Single-mode approaches do not provide necessary  accuracy for domain \\ \hline 
        Sensor consistency-based selection score, LiDAR guidance as semi-supervision for monocular detection \cite{hekimoglu2023multi, hekimoglu2024monocular} & 2024 &  KITTI, Waymo & Camera & Single-mode approaches do not provide necessary  accuracy for domain \\ \hline 
        \rowcolor{gray!10} 
        \textbf{Classification Entropy Querying, T-CRB Criteria \cite{ghita2023activelearning} (Ours)} & \textbf{2024} &\textbf{nuScenes, TUMTraf-I*} & \textbf{LiDAR+Camera, LiDAR} & Assumption of uniform label and point cloud distributions\\
     \hline\\[-8pt]
    \multicolumn{5}{l}{$^{\mathrm{*}}$TUMTraf-I Dataset.} \\
    \multicolumn{5}{l}{$^{\mathrm{\dag}}$Camera is used for pre-trained region proposals and is not included in the active learning loop.} \\
    \end{tabular*}
    
    \label{tab:al_det}
\end{table*}

\section{INTRODUCTION}

The annotation process remains a challenge, especially for 3D point clouds. According to \cite{song2015sun}, labeling a precise 3D box takes more than 100 seconds for an annotator. This scales up to a labeling time of 7,937 hours for the \textit{nuScenes} dataset (see Fig. \ref{fig:overview_figure}). Since deep learning models require a massive volume of labeled data, creating large-scale labeled 3D datasets is a challenge in developing robust 3D perception models. To alleviate this challenge, active learning aims to reduce the labeling costs by querying labels for a small fraction of unlabeled data, thus maximizing the model performance while annotating the fewest possible samples. While active learning significantly reduces the annotation costs by selecting the most informative image and point cloud frames, it concurrently elevates the computational expense due to the recurrent model training and evaluation cycles. Hence, active learning is especially useful toward continual learning methods that already facilitate these training recurrences, reducing the computational burden by facilitating more efficient model updates, thereby maintaining the gains of active learning without encountering prohibitive computational expenses.

With this work, we fill the research gap in using active learning for multi-modal 3D object detection. We adopt the work of \cite{luo2022exploring} as our baseline active learning framework and integrate it with the continuous training methods from \cite{dayoub2017episode}. Existing work just focused on a single modality \cite{hekimoglu2022efficient,villalonga2020co,ccatal2020anomaly,meng2021towards} or applied active learning for multi-task settings, like object detection and semantic segmentation \cite{hekimoglu2023multi}, but it is recognized that multi-modal approaches to detection are significantly stronger toward the safety-critical 3D object detection task.

\textbf{Our contributions are the following:}
\begin{itemize}
\item We propose the \textit{ActiveAnno3D} framework, which ensures that the selected data samples for labeling are of maximum informativeness to our model training.
\item We explore various continuous training methods and integrate the most efficient method regarding computational demand and detection performance.
\item We conduct extensive experiments of \textit{BEVFusion} \cite{liu2023bevfusion} and \textit{PV-RCNN} \cite{shi2020pv} on the \textit{nuScenes} \cite{caesar2020nuscenes} and \textit{TUM Traffic Intersection (TUMTraf-I)} dataset \cite{zimmer2023tumtraf}.
\item Finally, we integrate \textit{ActiveAnno3D} into the \textit{Providentia Annotation} (proAnno) platform to enable AI-assisted features and minimize manual labeling efforts.
\end{itemize}

\section{Related Research}

Active learning techniques can be broadly divided into two groups: those that depend on a particular model for a particular task and those that operate more generally on the data itself, independent of what problem the data will be solving (sometimes called ``diversity-based methods" \cite{liang2022exploring}). Though these techniques can be combined in creating a sampling strategy \cite{hekimoglu2022efficient}, the benefit of using a task- or model-dependent active learning strategy is that the selected data will be, in theory, optimal for improved performance of the desired task. In our research, we specifically consider the task of 3D object detection, which is critical for safety in autonomous driving. For this reason, we consider only these task-dependent, model-specific, uncertainty-driven approaches to active learning. We frame this as an episodic active learning task \cite{dayoub2017episode}, where some collection of data is gathered. Once a certain accumulation of data has been made, the active learning algorithm re-samples the available pool and uses an acquisition function to select samples to be labeled. While general active learning is well-researched on demonstration ML benchmark datasets, in the following sections, we describe the unique challenges associated with active learning in the domain of real-world autonomous driving.

\begin{figure*}
    \centering
    \includegraphics[width=.99\textwidth]{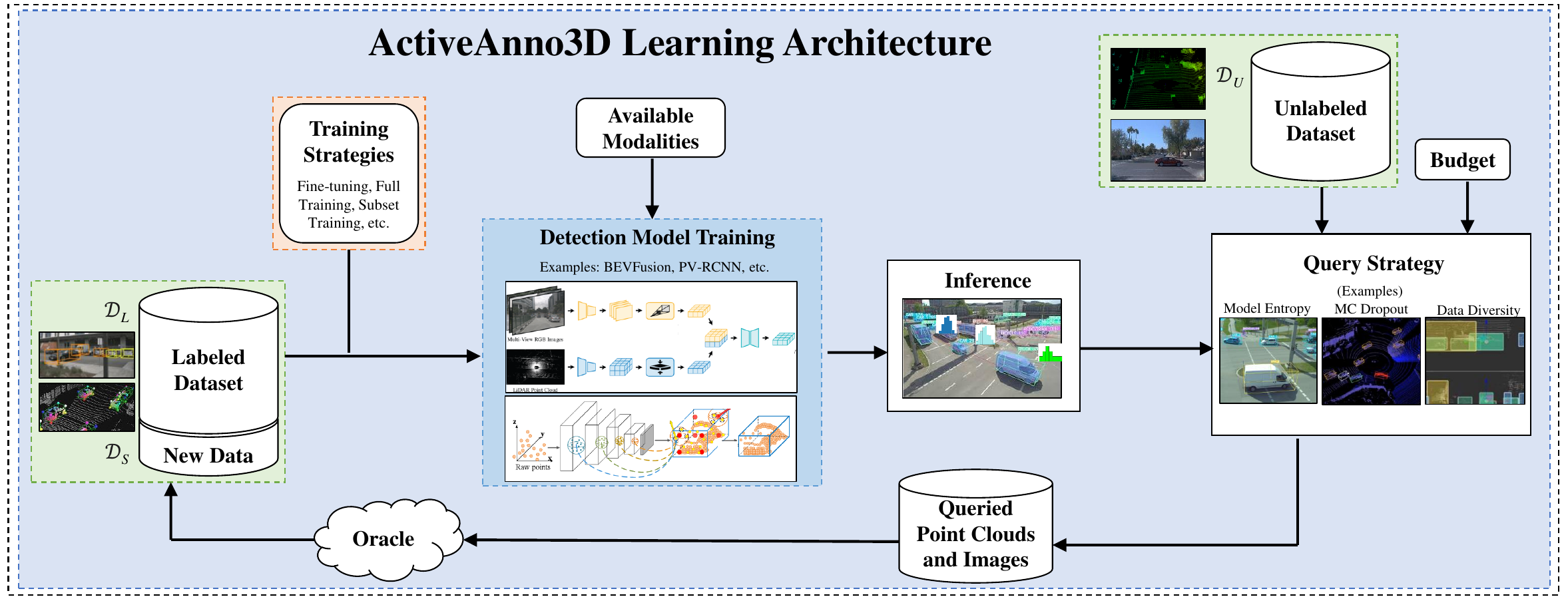}
    \caption{The generalized active learning flow involves the selection of data from an unlabeled pool according to an acquisition function, which, in the case of uncertainty-driven AL, utilizes the trained model or, in the case of diversity-driven AL, may be independent of the training. This selected data is then annotated by an oracle and aggregated with previously labeled data. Whether or not all data or just the new data is used in the next training step is determined by the choice of training strategy. The variety of possible acquisition and training techniques and unique domain challenges posed by autonomous driving make active learning an opportune environment for innovation toward safe and accurate learning.}
    \label{fig:enter-label}
\end{figure*}

Active Learning for Autonomous Driving must be:
\subsubsection{Multitask}
Identifying data optimal for one task is a good step. Still, at a practical level, if the purpose of active learning is to reduce the annotation budget, this does little good for an autonomous system that must use the same carefully curated and sampled data to solve multiple tasks---trajectory prediction \cite{greer2021trajectory, lefevre2015learning}, sign and light detection and classification \cite{salient, greer2023salient, greer2023robust}, dynamic object detection and tracking \cite{zimmer2022survey, zimmer2023real, chiu2021probabilistic, zimmer2023infradet3d}, lane detection \cite{huang2021real}, vehicle landmark identification \cite{GREER2024}, and more---simultaneously. Fortunately, multi-task consistency measurements can inform sampling in a way that considers the performance of multiple goal tasks \cite{xiao2020efficient, hekimoglu2023multi}. 

\subsubsection{Multi-modal}
While autonomous vehicle testbeds may be configured with a variety of sensors, the state of the art in 3D object detection, critical for collision-free driving, utilizes both LiDAR (for accurate ranging) and image (for semantic utility). At the time of this writing, the top 33 models on the \textit{nuScenes} 3D object detection leaderboard utilize both modalities (in fact, with two in the middle of the pack, further including radar). Most active learning research for 3D detection does not include the visual modality, and those that do leave their image-related modules out of the learning loop. We induce active learning on a state-of-the-art LiDAR+Camera architecture.

\subsubsection{Open-set}
We also note that data in the realm of autonomous driving is fundamentally open-set, meaning that the items labeled in a dataset today may not fully represent all items that might appear on the road when driving in the real environment. Standard active learning methods cannot perform as effectively in the open-set setting due to a class distribution mismatch between labeled and unlabeled data \cite{safaei2023entropic}, which also makes a strong case for the importance in the classification stage of an active learning model, as uncertainty in classification provides a means toward the identification of open-set class instances. Safaei et al. \cite{safaei2023entropic} show that using a closed-set entropy on class probabilities and distance entropy on learned features can be an effective learning strategy to efficiently learn on an unknown open-set pool, though this is yet to be applied to autonomous driving data, as their experimentation is done over the \textit{CIFAR} and \textit{TinyImageNet} benchmarks. This further motivates our exploration of entropy-based active learning within the autonomous driving domain.

Related to the open-set environment, though driving environments may differ regionally, there are fundamental similarities in the 3D detection problem that may facilitate efficient cross-domain knowledge transfer for learned models. Active learning strategies, and more specifically, active domain adaptation, are useful in improving learning for a target domain by sampling from a different source domain \cite{yuan2023bi3d}. However, these techniques have currently been evaluated only in the LiDAR learning setting, and there is room for multi-modal adaptation to improve accuracy. 

In Table \ref{tab:al_det}, we present an approximate chronological view of techniques that have been researched and evaluated on autonomous driving-related datasets for 3D object detection. The first active learning application to 3D object detection \cite{feng2019deep}, evaluated on the KITTI dataset, explored uncertainty-based techniques of Monte-Carlo dropout and deep ensembles; since then, the state of the art in 3D object detection has advanced, and our research provides further evidence that active learning methods which draw from model uncertainty are still relevant, even with updated detection methods and on a further variety of driving-related datasets. In developing active learning techniques that are particular for 3D object detection, Luo et al. \cite{luo2022exploring} define three mathematical terms (``CRB") to determine samples: 
\begin{enumerate}
    \item \textit{Label \textbf{C}onciseness}: aiming for the distribution of sampled labels to resemble the uniform distribution,
    \item \textit{Bounding box \textbf{R}epresentativeness}: aiming to select samples which the model would learn similar patterns for as reflected in the model gradients, and
    \item \textit{Geometric \textbf{B}alance}: aiming to learn from samples that fail to match expected point densities relative to an object's distance from the sensor.
\end{enumerate}   
In our empirical study, we examine the benefits of active learning on two datasets using two methods: one which acquires LiDAR samples according to these criteria of conciseness, representativeness, and geometric balance, and one which acquires combined LiDAR+Camera samples using object class entropy.

Our research represents the first investigation of active learning for 3D object detection (1) evaluating \textit{BEVFusion} on the \textit{nuScenes} dataset with an uncertainty-driven query strategy, and (2) evaluating \textit{PV-RCNN} using multiple query strategies on the roadside infrastructure \textit{TUM Traffic Intersection} dataset \cite{zimmer2023tumtraf}.

\section{Methods}

\subsection{Problem Statement}

Consider the problem of 3D object detection from \textbf{orderless LiDAR point clouds} represented as \( \mathcal{P} = \{ (x, y, z, \varepsilon) \} \), where \( (x, y, z) \) denotes the 3D location and \( \varepsilon \) represents the reflectance, and  \textbf{synchronous images} represented as \( \mathcal{I}_v(r, c) \), where \( \mathcal{I}_v(r, c) = \{ R, G, B\} \) is a 3-channel image of dimensions $r\times c$ from view $v$, and the collection of images from all views $v$ is referred to as \( \mathcal{I} \).   The objective is to localize objects of interest in the form of a set of 3D bounding boxes.

Let \( \Theta = \{ \mathbf{b}_k \}_{k \in [N_B]} \) be the set of 3D bounding boxes, where \( N_B \) is the number of detected bounding boxes. Each bounding box \( \mathbf{b}_k \) is associated with a label \( \mathbf{y}_k \) from a set \( \mathcal{Y} = \{1, \ldots, C\} \), where \( C \) is the number of classes to predict.

The parameters of each bounding box \( \mathbf{b}_k \) are given by \( (p_x, p_y, p_z, l, w, h, \theta) \), representing the coordinates of the box's center (\(p_x, p_y, p_z\)), length (\(l\)), width (\(w\)), height (\(h\)), and orientation angle (\(\theta\)).

At the training phase, a small set of labeled point clouds and associated images \( \mathcal{D}_L = \{(\mathcal{P}, \mathcal{I}, \mathbf{B}, \mathbf{Y})_i\}_{i \in [m]}\) is provided, where \( m \) is the number of labeled point clouds. Additionally, an unlabeled set of point clouds and associated images \( \mathcal{D}_U = \{ \mathcal{P}_j, \mathcal{I}_j \}_{j \in [n]} \) is available.

In each active learning round \( t \), guided by the active learning policy, a subset of the unlabeled raw data is selected. Subsequently, the labels of the 3D bounding boxes for the selected subset are queried from an oracle to construct \( \mathcal{D}_{\text{query}, t} \). The 3D object detection model is initially trained using \( \mathcal{D}_L \), then trained by \( \mathcal{D}_L \) and \( \mathcal{D}_{\text{query}, t}\) for all following rounds $t$ until the selected samples reach the final budget.

\begin{table}
    \caption{This table shows the mAP score for the random sampling baseline and the entropy querying method using the LiDAR-only \textit{PV-RCNN} model trained on the \textit{TUM Traffic Intersection} dataset, and the camera+LiDAR \textit{BEVFusion} trained on \textit{nuScenes}. The results are compared to the respective 100\% data usage results.}
    \resizebox{\linewidth}{!}{
    \setlength{\tabcolsep}{3pt}
    \begin{NiceTabular}{@{\extracolsep{4pt}}cccccc@{\extracolsep{0pt}}}
        \CodeBefore
        \rectanglecolor{gray!10}{11-1}{11-6}
        \Body
        \hline
        \multicolumn{2}{c}{\textbf{Labeled Pool}} & \multicolumn{2}{c}{\textbf{LiDAR-Only (PV-RCNN)}} & \multicolumn{2}{c}{\textbf{LiDAR+Camera (BEVFusion)}} \\
        \cmidrule{3-4}
        \cmidrule{5-6}
        \setlength{\tabcolsep}{0pt}
        Round & \% & Random & Entropy & Random & Entropy \\
        \hline \hline 
        \rule{0pt}{2ex} 1  & 10 & 51.03 & 54.32 (\scriptsize{\color{ForestGreen}{+3.29}}) & 30.95 & 31.06 (\scriptsize{\color{ForestGreen}{+0.11}})\\
        \rule{0pt}{2ex} 2 & 15 & 61.98 & 62.24 (\scriptsize{\color{ForestGreen}{+0.26}}) & 34.19 & 36.39 (\scriptsize{\color{ForestGreen}{+2.20}})\\
        \rule{0pt}{2ex} 3 & 20 & 69.84 & 68.23 (\scriptsize{\color{Red}{-1.61}}) & 38.00 & 40.41 (\scriptsize{\color{ForestGreen}{+2.41}})\\
        \rule{0pt}{2ex} 4 & 25 & 74.82 & 72.40 (\scriptsize{\color{Red}{-2.42}})& 42.36 & 42.17 (\scriptsize{\color{Red}{-0.19}})\\
        \rule{0pt}{2ex} 5 & 30 & 77.25 & \underline{76.56} (\scriptsize{\color{Red}{-0.69}})& \textbf{44.94} & 45.57 (\scriptsize{\color{ForestGreen}{+0.63}})\\
        \rule{0pt}{2ex} 6 & 35 & 75.40 & 75.00 (\scriptsize{\color{Red}{-0.40}}) & \underline{44.74} & 46.76 (\scriptsize{\color{ForestGreen}{+2.02}})\\
        \rule{0pt}{2ex} 7 & 40 & \underline{77.03} & 75.48 (\scriptsize{\color{Red}{-1.55}})& - & \underline{49.24}~~~~~~~~ \\
        \rule{0pt}{2ex} 8 & 50 & \textbf{79.09} & \textbf{77.25} (\scriptsize{\color{Red}{-1.84}})& - & \textbf{64.31}~~~~~~~~~\\
        \hline
        \rule{0pt}{2ex} \textbf{SOA (No AL)} & \textbf{100} & \multicolumn{2}{c}{\textbf{83.50}} & \multicolumn{2}{c}{\textbf{75.00}} \\
        \hline
    \end{NiceTabular}
    }
    \label{tab:AL_results}
\end{table}

\subsection{Active Learning for 3D Object Detection}
3D object detection is crucial for autonomous driving and mobile robotics, yet building precise 3D object detectors relies on the availability of extensive labeled datasets, a task that is particularly challenging under constrained annotation budgets. Active learning provides a promising solution to reduce labeling costs by querying labels for only a small subset of the unlabeled data pool. A criterion-based query selection process iteratively selects the most informative samples for the subsequent model training until the labeling budget is exhausted. Building on this, we deployed a pool-based active learning framework to a two-stage LiDAR-only 3D object detector and a multi-modal 3D object detector from the entropy querying strategies in \cite{greer2024why}. We examine the use of continuous training strategies to address the resource-consuming nature of active learning.

\subsection{Entropy Querying}
The query strategy, also known as the acquisition function, is the most critical component of the active learning framework as it identifies the most informative data samples according to a pre-defined criterion, and queries labels from the human annotator. We employ entropy querying that depends on the model's uncertainty in predictions. The model's uncertainty is measured using the entropy formula:
\begin{equation}
H(X) = -\sum_{i}(P(x_i) \log P(x_i))
\end{equation}
This equation predicts the entropy \(H(x)\), where \(P(x)\) is the probability of the model predicting class \(x_i\).

\begin{figure*}
    \centering
    \includegraphics[width=.52\linewidth]{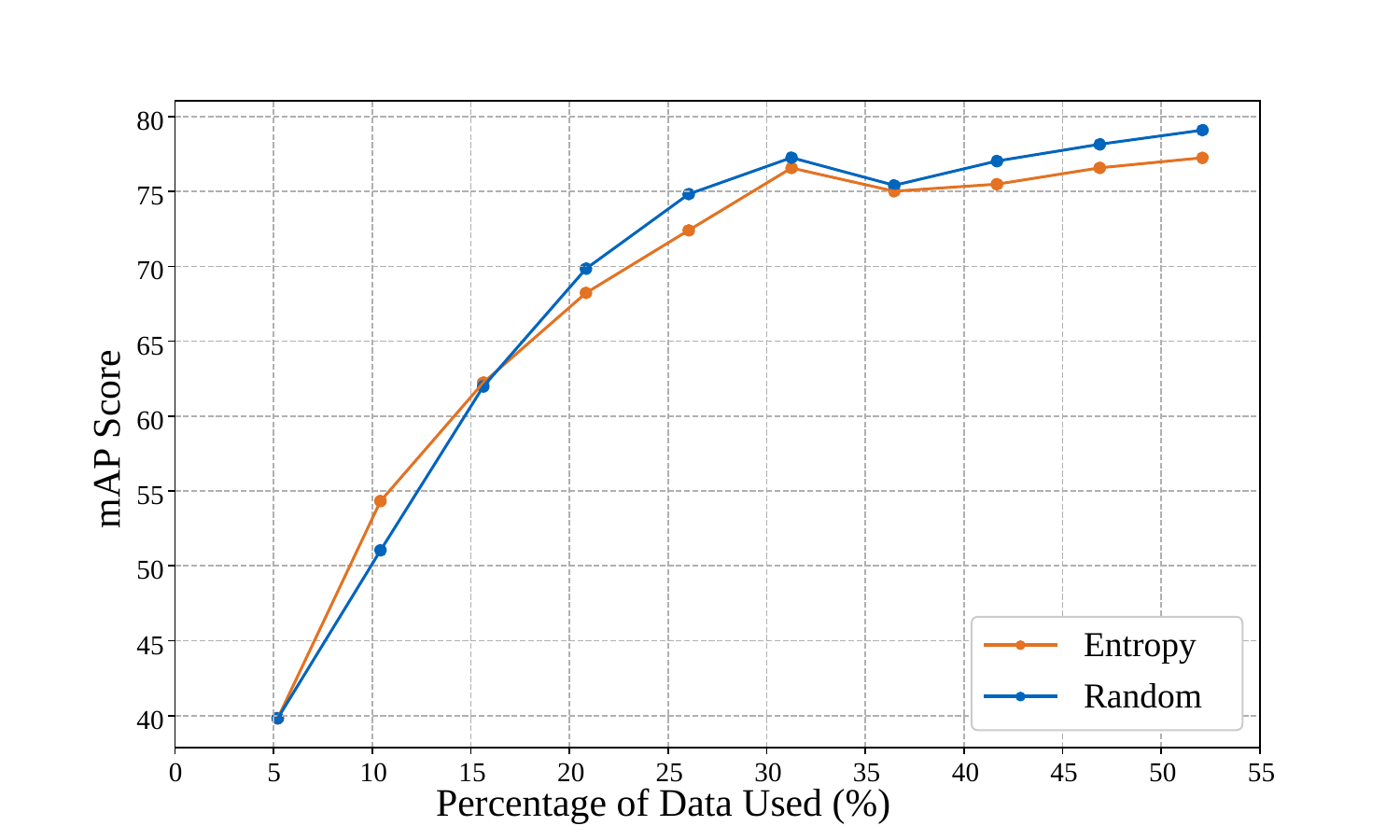}
    \includegraphics[width=.44\linewidth]{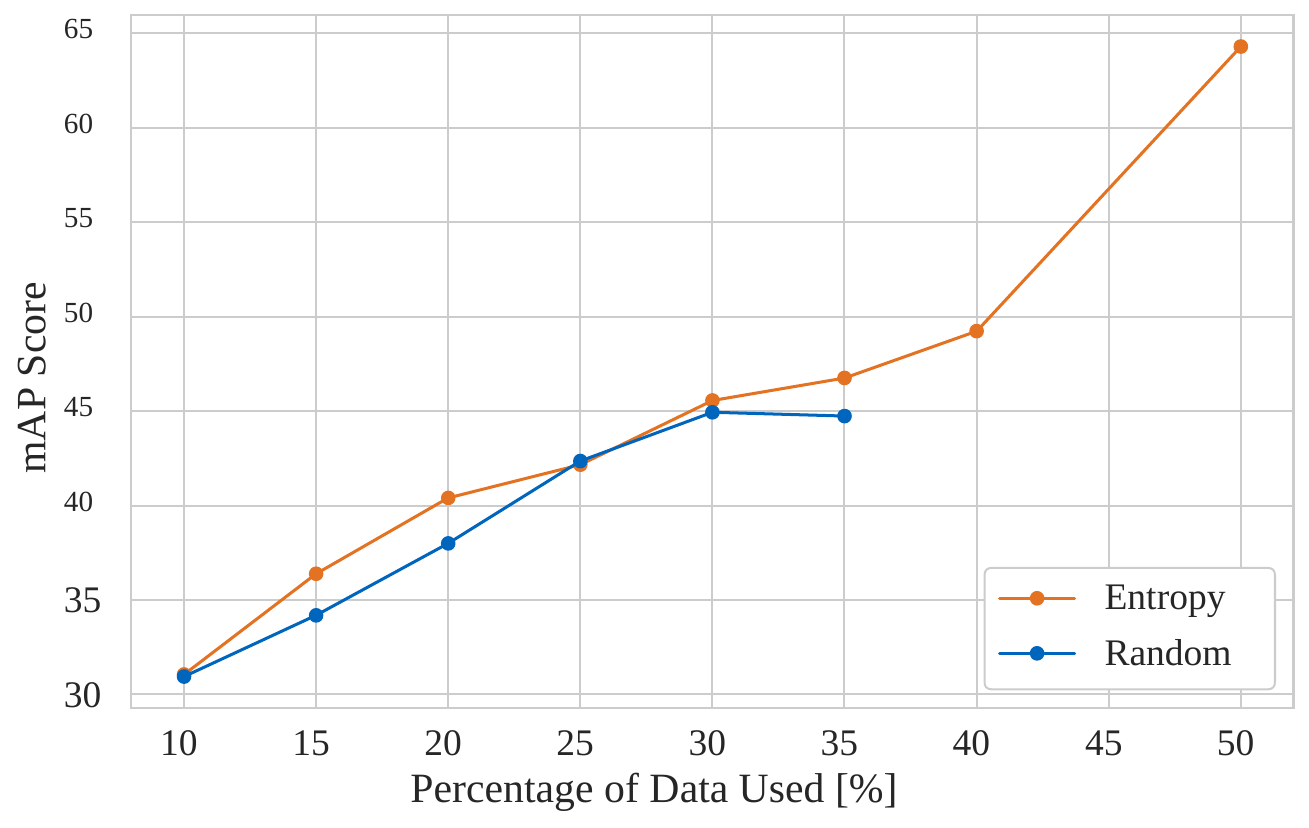}
    \caption{The graph on left illustrates the mAP scores achieved by the \textit{PV-RCNN} model on the \textit{TUM Traffic Intersection} dataset relative to the expanding size of the training set in the active learning setting with random and entropy queries separately. Similarly, the graph on right illustrates the mAP score achieved by the \textit{BEVFusion} model on the \textit{nuScenes} dataset relative to the expanding size of the training set.}
    \label{fig:active3d_results}
\end{figure*}

\subsection{Active Learning for LiDAR-only 3D Object Detection}
Initially, we employ the LiDAR-only model for 3D object detection known as the two-stage PV-RCNN \cite{shi2020pv}. Subsequently, during the inference stage, we utilize eight distinct query strategies, encompassing uncertainty-based, diversity-based, and hybrid methods. Each strategy is employed separately for selecting the most informative samples, guided by their individual selection criteria. Moreover, we introduce a temporal modification to the AL-baseline CRB \cite{crb}, which we refer to as T-CRB. T-CRB \cite{ghita2023activelearning} is designed to select the most informative temporally-consistent set of point clouds from the unlabeled data pool. This is particularly advantageous in scenarios involving point cloud labeling, where annotators seek to label a sequence of frames sequentially, aiming to maximize their contribution to the model's performance.

\subsection{Active Learning for Multi-Modal 3D Object Detection}

Second, we explore the BEVFusion model for 3D object detection, as outlined in the work by Liu et al. \cite{liu2023bevfusion}. They employ a multi-modal model combining camera and LiDAR data. This modality is relevant to exploring and comparing the results of our LiDAR-only approach. This approach is critical because, despite the existence of various techniques for achieving a unified representation of image and LiDAR data, LiDAR-to-Camera projections, as well as camera-only methods, introduce significant geometric distortions, and camera-to-LiDAR projections and LiDAR-only methods also face challenges in semantic-orientation tasks. BEVFusion is designed to create a unified representation that preserves geometric structure and semantic density.

The Swin-Transformer \cite{liu2021swin} serves as the image backbone, while VoxelNet \cite{zhou2018voxelnet} acts as the LiDAR backbone. To generate bird’s-eye-view (BEV) features for images, we employ a Feature Pyramid Network (FPN) \cite{lin2017feature} to fuse multi-scale camera features, resulting in a feature map one-eighth of the original size. Subsequently, images are down-sampled to 256x704, and LiDAR point clouds are voxelized to 0.075 m to obtain the BEV features necessary for object detection. These two modalities are fused using a convolution-based BEV encoder with residual blocks, mitigating local misalignment between LiDAR-BEV features and camera-BEV features. Especially from the uncertainty created in the depth estimation from the camera encoder, as a result, is not compared with ground truth \textit{i.e.,} this is a self-supervised learner. 

We then follow the same Active Learning scheme previously used for the LiDAR-only model.

\subsection{Continuous Training Strategies}
The framework illustrated in Figure \ref{fig:enter-label} follows an active learning approach, wherein an initial 3D detector undergoes pre-training on an initially labeled dataset. The query strategy identifies the most informative data samples through successive training episodes and solicits labels from an oracle. Subsequently, the initial model is updated using the augmented training set. However, this process is characterized by its time and resource intensiveness, as the model undergoes retraining from scratch in each episode on the expanding set of labeled data. Additionally, this approach neglects the knowledge accumulated in previous episodes. To address this challenge, we enhance the active training cycle by incorporating multiple continuous training strategies, as outlined in \cite{episodicAL}.

\section{Experiments}
We conduct a set of two experiments, integrating entropy-driven active learning over the multi-modal \textit{BEVFusion} model on two 3D object detection datasets, \textit{nuScenes}, and \textit{TUM Traffic Intersection}, and the two-stage \textit{PV-RCNN} model on the \textit{TUM Traffic Intersection} dataset.

\subsection{Datasets}
Large-scale datasets are essential for meaningful experiments. Therefore, we conducted our experiments on \textit{nuScenes} \cite{caesar2020nuscenes} and the \textit{TUM Trafic Intersection} dataset \cite{zimmer2023tumtraf}. The popular \textit{nuScenes} dataset provides 400k LiDAR point clouds and 1.4M images from which the annotators carefully labeled 40k frames. The dataset covers complex traffic scenarios, for example, at intersections or construction sites, and groups traffic participants into 23 different object classes. At the time of publication, \textit{nuScenes} was the first dataset that included night and rain scenarios. The data has been recorded by six cameras and one LiDAR (Velodyne HDL32E) mounted on a vehicle. \textit{nuScenes} contains 1000 scenes, 700 training scenes, 150 validation scenes, and 150 test scenes. We start with an initial split of 100 training scenes and iteratively add 50 scenes for each subsequent active learning round until 35\% of the data is used resulting in six training rounds.


As mentioned above, \textit{nuScenes} only covers the vehicle perspective. Therefore, we also used the \textit{TUM Traffic Intersection} dataset to experiment with further perspectives, considering the roadside perspective from a gantry bridge at a complex intersection. Two LiDARs (Ouster OS1-64, gen. 2) mounted at a height of 7 m on the \textit{Providentia++} test field \cite{kraemmer2022providentia} near Munich generated the relevant data for this work. The recorded traffic scenarios cover complex maneuvers (left/right turns, overtaking, U-turns), with 25\% at nighttime, including heavy rain, and 75\% during the day. In total, the dataset contains, 4,800 point clouds with more than 57.4k manually high-quality labeled 3D boxes categorized in ten object classes, labeled using the \textit{Providentia Annotation} \textit{proAnno} framework, an improved version of 3D BAT \cite{zimmer20193d}. After point cloud registration, the dataset contains 2,400 registered point clouds and 38,045 registered 3D boxes, which we used for this work. We split the data into training (80\%), validation (10\% ), and test (10\%)  sets. Here, we performed our active learning experiments on the train set with 1,920 samples. The initial labeled set $D_{pre}$ contained approximately 5\% (100 samples). We selected an additional 100 samples every training episode until the labeling budged (50\% of the training data) was exhausted. Similar to our \textit{nuScenes} experiments, we trained an oracle model on the complete training data of the \textit{TUM Traffic Intersection} dataset to set a baseline for measuring the effectiveness of our active learning approach.


\subsection{Results}

\begin{figure}
    \centering
    \includegraphics[width=1\linewidth,trim={1.5cm 0 2cm 1.7cm},clip]{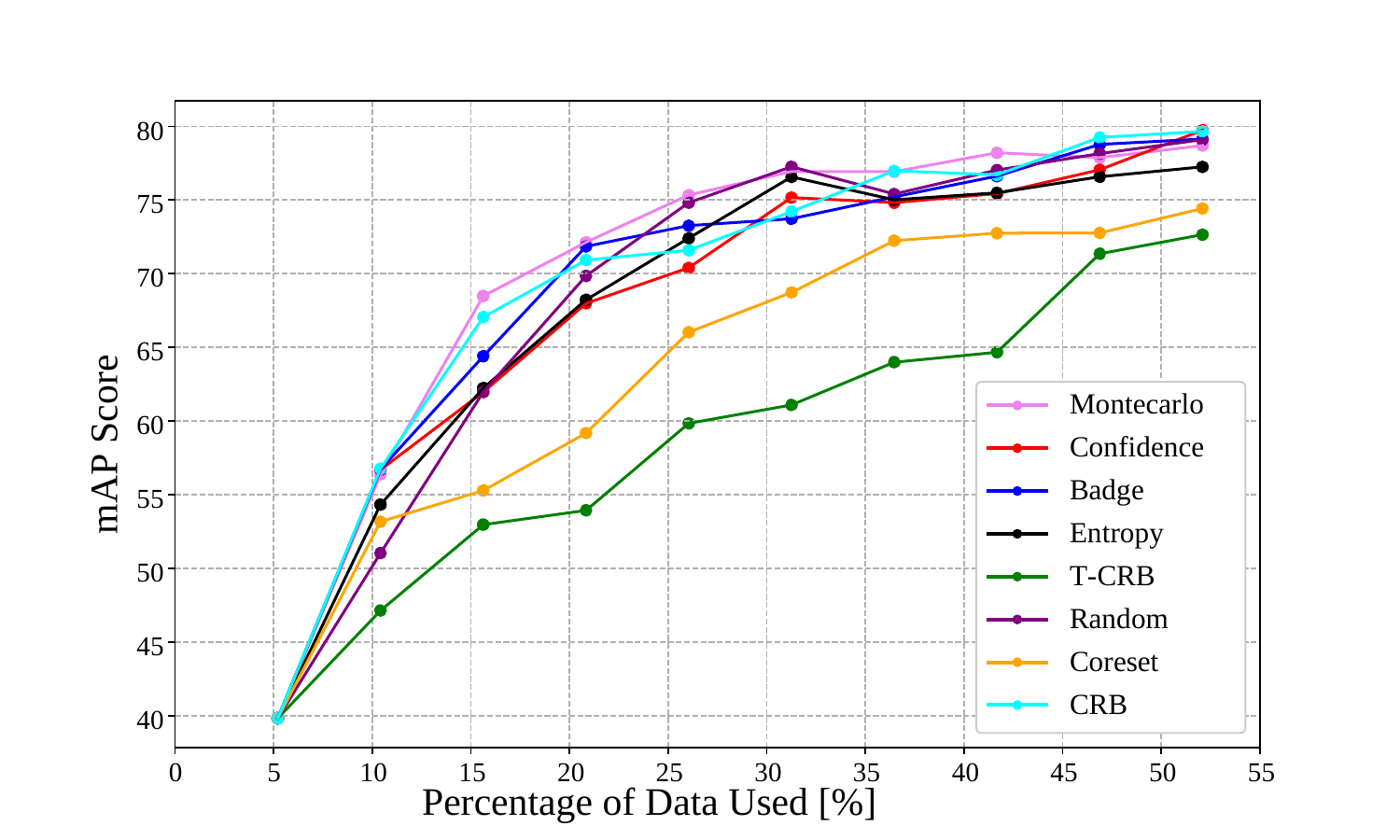}
    \caption{The graph illustrates the mAP scores achieved by the \textit{PV-RCNN} model on the \textit{TUM Traffic Intersection} dataset relative to the expanding size of the training set in the active learning setting with different query strategies.}
    \label{fig:active3d_results_all}
\end{figure}

Figure \ref{fig:active3d_results_all} provides an insight into the performance linked with various query strategies within active learning for 3D object detection. The evaluation encompasses a variety of strategies such as \textit{BADGE} \cite{badge}, \textit{CoreSet} \cite{coreset}, \textit{Monte-Carlo} sampling, confidence sampling, \textit{CRB} \cite{crb}, and the temporally-modified \textit{CRB} (\textit{T-CRB}). Such analysis is useful towards pinpointing the most effective approach tailored to the unique characteristics of the \textit{TUM Traffic Intersection} dataset.
The results presented in Table \ref{tab:AL_results} show that the \textit{PV-RCNN} detector using the random sampling query strategy trained on the \textit{TUM Traffic Intersection} dataset, outperformed the entropy query strategy as illustrated in the left of Figure \ref{fig:active3d_results}. However, this pattern does not persist in the LiDAR+camera model. As observed in right of Figure \ref{fig:active3d_results}, in every round except round four, entropy querying outperforms random querying.


\begin{figure*}[t]
\centering
\minipage{0.25\textwidth}
  \includegraphics[width=\linewidth,trim={0 6.6cm 0 0},clip,frame]{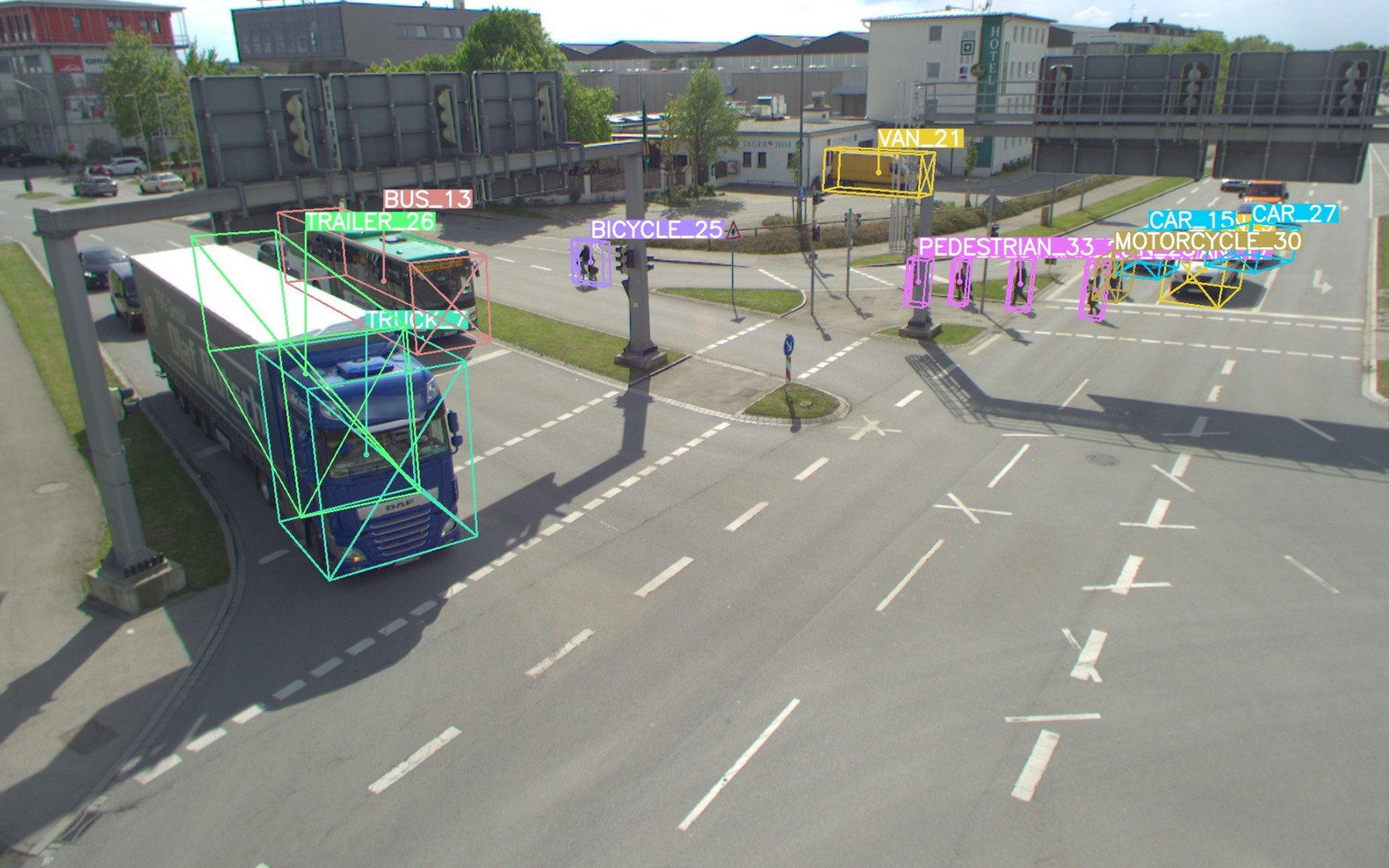}
  \caption*{\footnotesize{CRB}}
\endminipage
\minipage{0.25\textwidth}
  \includegraphics[width=\linewidth,trim={0 6.6cm 0 0},clip,frame]{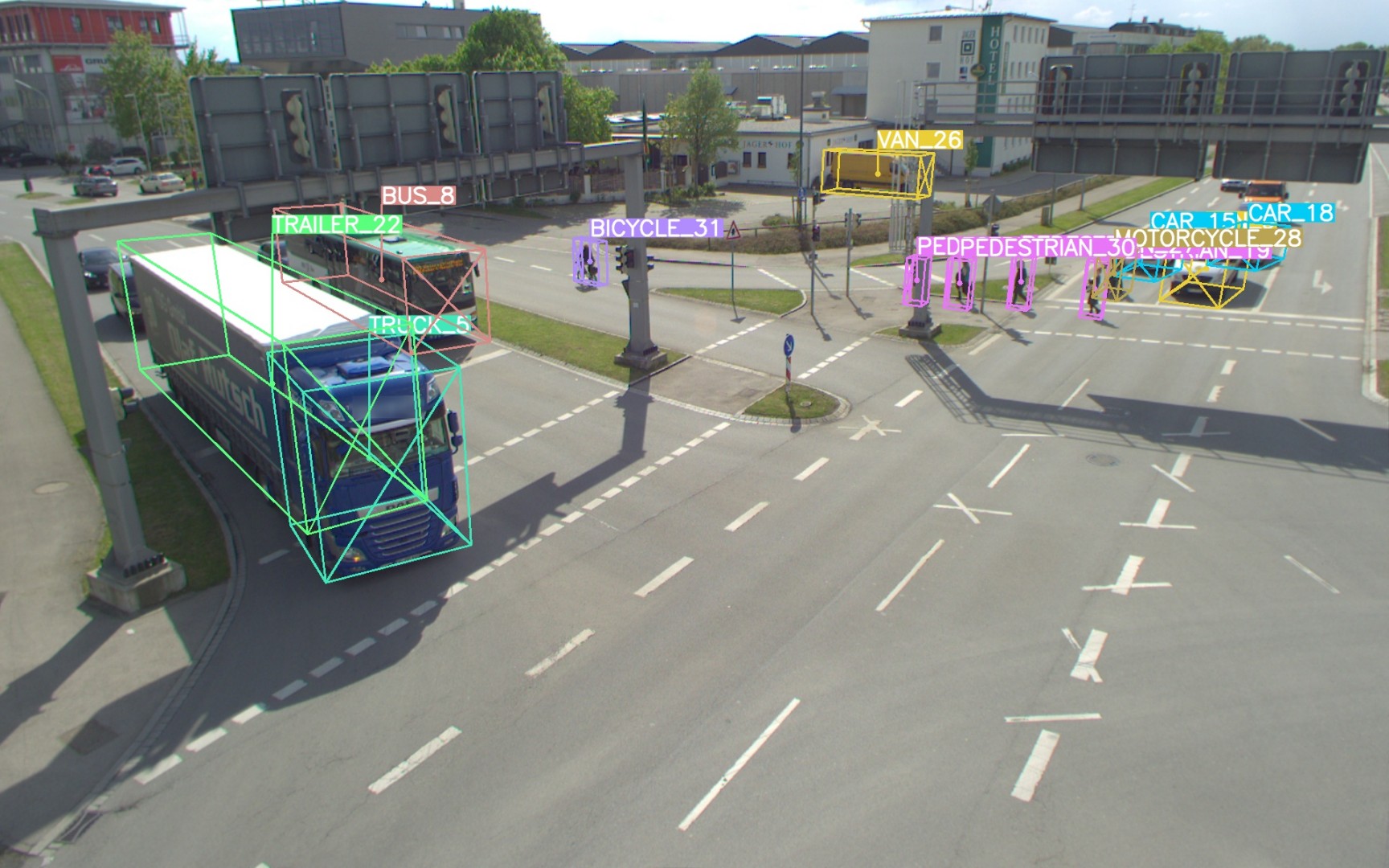}
  \caption*{\footnotesize{Oracle}}
\endminipage
\minipage{0.25\textwidth}
  \includegraphics[width=\linewidth,trim={0 1.0cm 0 0},clip,frame]{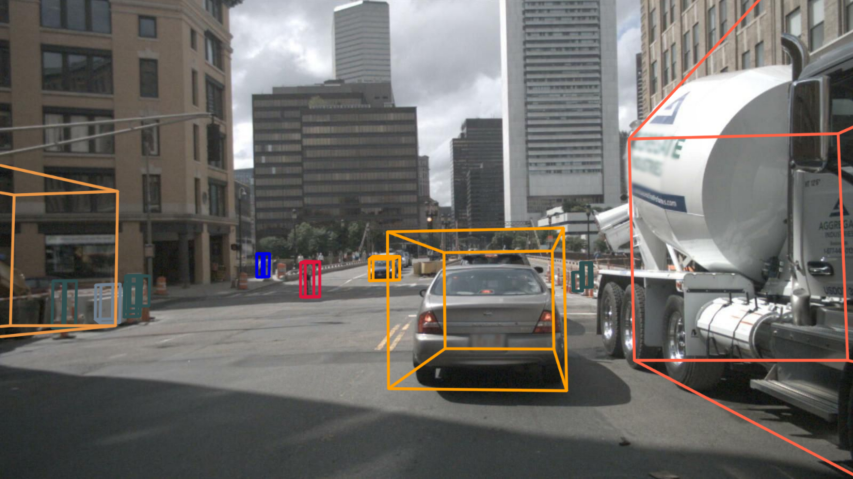}
  \caption*{\footnotesize{Entropy}}
\endminipage
\minipage{0.25\textwidth}%
  \includegraphics[width=\linewidth,trim={0 1.0cm 0 0},clip,frame]{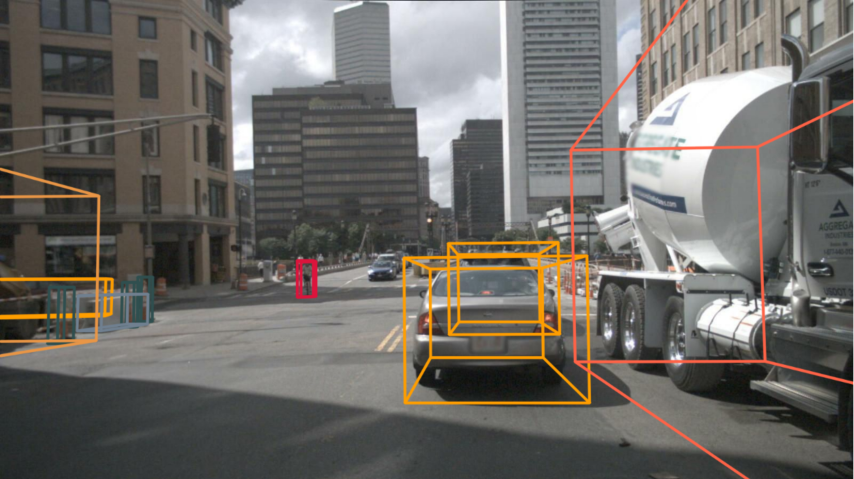}
  \caption*{\footnotesize{Oracle}}
\endminipage
\caption{Qualitative results are illustrated by two pairs of images. The left pair is from the TUM Traffic Intersection dataset, and the right pair is from nuScenes. For each pair, the left image shows the predicted labels for each class, with each class represented by a different color. The right image of each pair shows the predictions made by learning on the complete dataset. Both results are quite similar, showing the efficiency of the active learning technique.}
\label{fig:qualitative_results} 
\end{figure*}

For the qualitative analysis, we compare a pair of images from each dataset, with the predicted annotations and the corresponding ground truth.
For the \textit{TUM Traffic Intersection} dataset, illustrated in Figure \ref{fig:qualitative_results}, it can be seen that all the eight classes are present in the image, and the corresponding bounding boxes reflect the data well, as confirmed by the green bounding boxes seen in the left image.


\setlength{\fboxsep}{0pt}%
\setlength{\fboxrule}{1pt}%

For the \textit{nuScenes} dataset illustrated in Figure \ref{fig:qualitative_results} there are also many classes present, though not all. On the right image, it is also possible to see that all predictions are correct, except for one instance in the back where an object is classified as a pedestrian, whereas in the image showing predictions made by the model trained on the full dataset, there is no annotation to represent this object.


\subsection{Discussion}
The variability in performance between datasets, where entropy querying outperformed random querying in the scenario of \textit{BEVFusion} with \textit{nuScenes}, while random querying approximately matches entropy querying in \textit{PV-RCNN} with \textit{TUM Traffic Intersection}, highlights the importance of understanding the data characteristics and intricacies. Random sampling, for instance, characterized by its unselective and undiscriminating approach, often serves as a robust baseline. Its lack of bias towards specific data characteristics ensures a diverse and representative sample of the scenario. This characteristic proves advantageous in scenarios with moderate-range objects, where the data achieves an equilibrium between density and sparsity, as the case in the TUMTraf Intersection Dataset. The ability of random sampling to provide a varied set of frames contributes to its effectiveness in such contexts. On the other hand, entropy sampling, designed to target the uncertainty of object classification, may introduce bias towards certain object classes. This bias has the potential to limit the selection of a diverse range of frames, which, in turn, might negatively impact the overall learning performance for certain datasets.

The unexpected superiority of random querying over entropy querying in this case highlights the nuanced relationship between the acquisition function and the dataset characteristics, emphasizing the need for careful consideration when choosing active learning strategies tailored for specific scenarios.
Moreover, strategies such as CRB, random sampling, and confidence sampling show high mAP scores, further highlighting the importance of strategy selection based on specific data characteristics in active learning settings. CRB achieves notable overall performance, attributed to its hierarchical and hybrid nature, which combines uncertainty-based criteria, diversity-based criteria, and geometric characteristics. This enables the query strategy to balance between diverse detection scenarios defined by distance categories. In contrast, confidence sampling, focusing on the uncertainty of objectness in frames, attains the highest mAP performance due to its narrower focus. This allows it to select frames that facilitate the model's learning about various objects. Unlike entropy sampling, which targets the uncertainty of object classification and might introduce a bias towards certain object classes, confidence sampling's focused approach contributes to its effectiveness in providing informative samples.

\section{CONCLUSIONS AND FUTURE WORK}
To overcome the high labeling costs, we have proposed an active learning framework termed \textit{ActiveAnno3D} for multi-modal 3D object detection. We provide a promising framework to reduce labeling costs by querying labels for only a small subset of the unlabeled data pool. A criterion-based query selection process iteratively selects the most informative samples for the subsequent model training until the labeling budget is exhausted. We adopt \cite{luo2022exploring} as our baseline active learning framework, deploying a hierarchical, cost-effective point cloud acquisition criteria at the 3D bounding box level, termed CRB. To prevent retraining the model from scratch every time new data is selected, we explore various continuous training methods and integrate the most efficient method regarding computational demand and detection performance. Finally, we show that our active learning framework achieves high performance (77.25 mAP compared to 83.50 mAP) with reduced data utilization (50\%) without placing excessive demands on computational resources.
The camera-LiDAR BEVFusion model achieved an mAP of 64.31 when using 50\% of the data for training that was selected with the entropy-based sampling strategy, compared to 75.0 mAP when using all the data.

In the future, we plan to improve the CRB query strategy. The current version operates under the assumption that the unlabeled data pool possesses both a uniform label distribution and a uniform point density distribution. By incorporating a Bayesian approach, benefiting from the iterative process of active training, and evolving the training set, it may become possible to mirror the test set. Another future direction will be guided active selection. The active query strategy can be tailored to suit the preferences of a human annotator. If specific classes need to be labeled, the strategy can be modified to evaluate the informativeness of these data samples. The framework will then return the most informative samples about the desired object classes. We also plan to make our ActiveAnno3D framework compatible with multiple baselines (e.g., BEVFusion) and ease the integration of new architectures. Finally, we plan to extend BEVFusion with different sampling strategies and compare it with the results of the LiDAR-only PV-RCNN method.

Overall, the demonstrated sensitivity of 3D object detection performance to active learning strategy is dependent on the nature of the available data, both offering a clear value in adopting active learning in general for efficient learning, and encouraging further research and analysis on optimal strategies for individual and multiple tasks in safe autonomous driving. 







\section*{ACKNOWLEDGMENT}
This research was supported by the Federal Ministry of Education and Research in Germany within the $\text{\textit{AUTOtech.agil}}$ project, Grant Number: 01IS22088U. The authors thank Qualcomm for the support of the Qualcomm Innovation Fellowship, mentors Per Siden and Varun Ravi for their feedback, and Mathias V. Andersen for assistance in algorithm implementations. 


\bibliographystyle{IEEEtran}
\bibliography{root.bib}

\end{document}